\documentclass[runningheads]{llncs}
\usepackage[T1]{fontenc}
\usepackage{graphicx}
\usepackage{makecell}
\usepackage{multirow}

\usepackage{booktabs}
\usepackage{tabularx}
\usepackage{multirow}
\usepackage{makecell}
\usepackage{array}
\usepackage[flushleft]{threeparttable}
\usepackage{cite}
\usepackage{placeins} 
\usepackage{url}

\usepackage[most]{tcolorbox}
\usepackage{xcolor}
\usepackage{enumitem}
\usepackage{tikz}

\definecolor{requestablue}{RGB}{230,245,255}
\definecolor{gptpink}{RGB}{255,235,240}
\definecolor{labelblue}{RGB}{0,102,153}
\definecolor{answerpurple}{RGB}{120,60,150}

\newtcolorbox{requestabox}{
  enhanced,
  breakable,
  colback=blue!6,
  colframe=blue!45!black,
  boxrule=0.6pt,
  arc=2pt,
  left=6pt,right=6pt,top=6pt,bottom=6pt,
  before skip=8pt,
  after skip=8pt
}

\newtcolorbox{baselinebox}{
  enhanced,
  breakable,
  colback=red!6,
  colframe=red!45!black,
  boxrule=0.6pt,
  arc=2pt,
  left=6pt,right=6pt,top=6pt,bottom=6pt,
  before skip=8pt,
  after skip=8pt
}

\newcommand{\qtype}[1]{%
  \tcbox[
    colback=labelblue!15,
    colframe=labelblue,
    arc=1.5mm,
    boxrule=0.4pt,
    left=3pt,
    right=3pt,
    top=1pt,
    bottom=1pt
  ]{\fontsize{8}{9}\selectfont\textbf{#1}}%
}

\newcommand{\answer}[1]{%
  \tcbox[
    colback=answerpurple!15,
    colframe=answerpurple,
    arc=1.5mm,
    boxrule=0.4pt,
    left=3pt,
    right=3pt,
    top=1pt,
    bottom=1pt
  ]{\fontsize{8}{9}\selectfont\textbf{Correct Answer: #1}}%
}

\newcommand{\qdotted}{%
  \par\vspace{1pt}%
  \noindent
  \begin{tikzpicture}
    \draw[dotted, line width=0.5pt, gray] (0,0) -- (\linewidth,0);
  \end{tikzpicture}%
  \par\vspace{1pt}%
}

\begin{document}
\title{Cognitively Diverse Multiple-Choice Question Generation: A Hybrid Multi-Agent Framework with Large Language Models}
\titlerunning{Hybrid Agentic Framework for Cognitively Diverse MCQ Generation}
%
%
\author{Yu Tian\inst{1} \and
Linh Huynh\inst{1} \and
Katerina Christhilf\inst{1} \and
Shubham Chakraborty\inst{1} \and
Micah Watanabe\inst{1} \and
Tracy Arner\inst{1} \and
Danielle McNamara\inst{1}}
\authorrunning{Tian et al.}
\institute{Arizona State University, 1151 S Forest Ave, Tempe, AZ 85287, USA\\
\email{\{ytian126, lthuynh1, kchristh, schak103, mwatana5, tarner, dsmcnama\}@asu.edu}
}
\maketitle              
\begin{abstract}
Recent advances in large language models (LLMs) have made automated multiple-choice question (MCQ) generation increasingly feasible; however, reliably producing items that satisfy controlled cognitive demands remains a challenge. To address this gap, we introduce ReQUESTA, a hybrid, multi-agent framework for generating cognitively diverse MCQs that systematically target text-based, inferential, and main idea comprehension. ReQUESTA decomposes MCQ authoring into specialized subtasks and coordinates LLM-powered agents with rule-based components to support planning, controlled generation, iterative evaluation, and post-processing. We evaluated the framework in a large-scale reading comprehension study using academic expository texts, comparing ReQUESTA-generated MCQs with those produced by a single-pass GPT-5 zero-shot baseline. Psychometric analyses of learner responses assessed item difficulty and discrimination, while expert raters evaluated question quality across multiple dimensions, including topic relevance and distractor quality. Results showed that ReQUESTA-generated items were consistently more challenging, more discriminative, and more strongly aligned with overall reading comprehension performance. Expert evaluations further indicated stronger alignment with central concepts and superior distractor linguistic consistency and semantic plausibility, particularly for inferential questions. These findings demonstrate that hybrid, agentic orchestration can systematically improve the reliability and controllability of LLM-based generation, highlighting workflow design as a key lever for structured artifact generation beyond single-pass prompting.

\end{abstract}
\section{Introduction}
Recent advances in generative artificial intelligence (AI), particularly large language models (LLMs), have enabled transformative applications across a wide range of domains, including education \cite{kasneci2023chatgpt, wang2024large}. State-of-the-art LLMs such as GPT-4 and GPT-5 are pretrained on massive, heterogeneous corpora and exhibit strong capabilities in language understanding and generation, often requiring little to no task-specific training through few-shot or zero-shot learning. These capabilities have opened new opportunities for automating complex language-mediated tasks that were previously labor-intensive, including question generation \cite{lee2024math,bulathwela2023scalable,mucciaccia2025automatic}, essay scoring \cite{lee2024applying, bui2025chatgpt}, and automated feedback generation \cite{dai2023can, liang2024can}.

Among these tasks, automated generation of multiple-choice questions (MCQs) represents a particularly challenging and practically important problem. MCQs are a widely used assessment format in which learners select the best possible answer from a set of alternatives \cite{butler2018multiple}. A typical MCQ consists of a question stem, a correct answer (key), and several distractors (plausible but incorrect options designed to differentiate levels of understanding). This structured format affords efficient administration and objective scoring. These properties of MCQs are critical to scalable deployment compared to other forms of assessment (e.g., essays). MCQs enable rapid feedback, consistent evaluation, and fine-grained item-level analysis, making them well suited for computer-based testing and learning analytics applications \cite{simkin2005multiple}.

From a systems engineering perspective, however, generating high-quality MCQs is not simply a matter of producing linguistically well-formed outputs. Effective MCQs must satisfy multiple structural and functional constraints simultaneously: the stem must clearly specify the task, the correct answer must be uniquely defensible given the source material, and distractors must be plausible, parallel in form, and semantically distinct \cite{moore2023developing}. Critically, MCQs may also differ in the reasoning demands they impose, ranging from questions that target explicitly stated information to those that require inference or synthesis across multiple ideas. We refer to this property as cognitive diversity, understood here as a property of generated artifacts that reflects differences in the reasoning processes required for successful response.

Automated MCQ generation has consequently attracted sustained attention from researchers seeking to reduce authoring burden while improving reliability and consistency \cite{mitkov2006computer,raina2022multiple, olney2023generating, pawar2024automated, mucciaccia2025automatic}. Advances in transformer-based LLMs have made it increasingly feasible to automatically generate high-quality MCQs. Empirical studies have shown that LLM-generated questions can approach human-authored items on several quality dimensions. For example, Olney \cite{olney2023generating} reported parity between LLM-generated and human-authored MCQs on most item-quality metrics, while Cheung et al. \cite{cheung2023chatgpt} found that ChatGPT-generated MCQs were broadly comparable to faculty-written items in terms of clarity, specificity, and suitability for graduate-level examinations.

Nevertheless, important challenges remain. In particular, reliably aligning generated questions with higher-order cognitive skills (e.g., inferential or analytical reasoning) continues to be a difficult task for LLMs. Recent evaluations that categorize questions by reasoning demand indicate that LLMs perform well for lower-level recall items, but performance degrades for questions targeting higher-order cognitive skills unless additional control mechanisms are introduced \cite{scaria2024automated}. This challenge has motivated research into hybrid approaches that augment LLMs with rule-based components, external knowledge sources, and structured control mechanisms, as well as agentic architectures that decompose complex generation tasks into coordinated subtasks \cite{li2023multi, hang2024mcqgen, bhowmick2023automating,mucciaccia2025automatic}.

Against this backdrop, we introduce ReQUESTA (\textbf{Re}ading \textbf{Q}uestion-generation \textbf{U}sing \textbf{E}ducational \textbf{S}mart \textbf{T}ext \textbf{A}gents), a hybrid multi-agent framework designed to generate MCQs with explicit control over cognitive focus. ReQUESTA integrates LLM-powered agents and rule-based components to generate three types of MCQs: (i) text-based (recall), (ii) inferential, and (iii) main idea (synthesis) questions. ReQUESTA targets reliability and controllability through orchestration, rather than relying on model scaling alone. The framework decomposes the MCQ authoring process into specialized subtasks—such as planning, question generation, evaluation, and formatting—each handled by a dedicated agent. By combining the generative strengths of LLMs with the precision and consistency of rule-based logic, ReQUESTA establishes a scalable and assessment-grounded workflow for automated MCQ generation.

ReQUESTA was first introduced in \cite{tian2026requesta}, which presented the core architecture and provided preliminary evidence that the ReQUESTA framework could reliably target discrete cognitive skills in generated questions. The present article substantially extends that work in both technical depth and empirical rigor. We provide a detailed account of ReQUESTA’s agentic design, including design rationales, agent interactions, and the division of labor between LLM-powered and rule-based components. We also report a multi-layered evaluation aligned with contemporary standards in educational measurement. Specifically, to test whether structured orchestration improves MCQ generation beyond single-pass prompting with the same base model, we compared items generated by ReQUESTA to those generated by a GPT-5 zero-shot baseline using psychometric analyses and expert judgments of item quality. Psychometric analyses examined item difficulty and discrimination, while expert evaluations focused on item clarity, construct relevance, and distractor quality. Our evaluations aimed to address the following research questions (RQs):

RQ1: Do ReQUESTA items exhibit stronger psychometric functioning (higher difficulty and discrimination) than a GPT-5 zero-shot baseline? 

RQ2: Does ReQUESTA improve expert-rated item quality, such as item clarity, topic relevance, and distractor construction (linguistic consistency, semantic plausibility, uniqueness)? 

While our prior work established proof of concept, the present study positions ReQUESTA as a mature, evaluation-driven framework for automated assessment generation. This study makes three primary contributions to research on LLM-based generation systems. First, it shows that meaningful improvements in automated MCQ generation can be achieved through workflow and orchestration design rather than model scaling, by holding the underlying LLM constant and introducing structured coordination across planning, generation, evaluation, and revision stages. Second, it proposes a controllable generation framework that combines the expressive power of LLMs with the reliability of deterministic, rule-based processing. Within this framework, agent boundaries, task specialization, and iterative feedback loops are explicitly engineered to regulate the reasoning demands and structural constraints of generated artifacts, moving beyond single-pass or monolithic prompting. Third, the study advances evaluation practice for structured generation systems by assessing system behavior using functional outcome measures (e.g., item difficulty and discrimination) rather than relying solely on surface-level indicators like grammaticality. Collectively, these contributions illustrate how agentic, workflow-level design can yield more reliable and controllable outcomes in LLM-based generation systems.

\section{Related Work}
\subsection{Classical Natural Language Processing (NLP) Approaches}
Automated MCQ generation has been an active area of research in educational technology and NLP for several decades. Over time, researchers have explored a wide range of computational approaches to automate different stages of the MCQ authoring process, with classical NLP techniques (e.g., keyword identification, semantic similarity scores) forming the foundation of much of this work. Although more recent systems increasingly rely on LLMs, many of the procedural assumptions and design principles established in classical NLP continue to inform modern hybrid frameworks, including the development of ReQUESTA.

Early approaches to automated MCQ generation typically decomposed the task into a sequence of subtasks: (1) text preprocessing, (2) sentence selection or content extraction, (3) stem and correct-answer generation, (4) distractor generation, and (5) post-processing steps such as verbalization, ranking, or filtering of candidate questions \cite{alsubait2015, kurdi2020systematic, madri2023comprehensive}. This explicit decomposition reflects the recognition that MCQ authoring is inherently multi-stage and that different subtasks benefit from different computational strategies. While advances in statistical and transformer-based NLP have substantially improved the mechanisms to implement these steps, task decomposition remains a widely adopted organizational strategy in contemporary MCQ generation systems \cite{kurdi2020systematic, mucciaccia2025automatic}.

Many classical systems relied heavily on syntactic analysis of the input text, using features such as part-of-speech tags, dependency relations, or parse trees to identify key phrases, extract candidate answers, and transform declarative sentences into interrogative forms \cite{mitkov2003computer, kalady2010natural}. Syntactic similarity was also frequently leveraged to support distractor generation and selection, such as by identifying structurally similar phrases or entities that could serve as plausible incorrect options \cite{aldabe2009study}. Although such methods were limited in their ability to capture deeper semantic relationships, they established important principles regarding structural parallelism and stylistic consistency among answer options. These principles were often preserved in modern automated MCQ generation systems, particularly in efforts to improve distractor quality and reduce superficial cues to the correct answer \cite{kurdi2020systematic, moore2023developing}.

Subsequent research also introduced semantics-based approaches that aimed to move beyond lexical and syntactic manipulation toward deeper representations of meaning. These methods leveraged techniques such as semantic role labeling to identify predicates, arguments, and semantic roles within sentences, enabling more principled extraction of question-relevant content \cite{fattoh2014sematic, mannem2010question, araki2016generating}. While classical semantic models were often limited by coverage and domain specificity, they highlighted the importance of explicitly modeling meaning and relationships when generating assessment items. This emphasis on semantic coherence, particularly for identifying central concepts and constructing plausible distractors, continues to inform contemporary system-level approaches to automated MCQ generation \cite{mitkov2006computer, kurdi2020systematic}.

\subsection{Transformer and LLM-based Approaches}
In recent years, transformer-based pretrained language models have substantially reshaped the landscape of automated question generation. Compared with earlier rule-based and feature-driven NLP pipelines, transformer models can condition on longer spans of context and generate fluent, well-formed questions with minimal manual engineering \cite{akyon2022automated, goyal2024automated}. This capability has made high-quality MCQ generation increasingly feasible across domains, particularly when source content is provided as extended passages or textbook excerpts rather than isolated sentences.  

The emergence of large, instruction-tuned language models has further enabled prompt-based MCQ generation without explicit task-specific training. In these approaches, LLMs are prompted to generate questions directly from instructional materials, often relying on carefully designed prompts to guide output structure and content. Empirical evaluations suggest that, under controlled conditions, such models can produce MCQs comparable to those authored by human experts on several quality dimensions. Olney \cite{olney2023generating}, for instance, reported that LLM-generated MCQs derived from textbook content matched human-authored items on most evaluated metrics. In a multinational study of medical education, Cheung et al. \cite{cheung2023chatgpt} found that ChatGPT-generated MCQs exhibited levels of clarity, specificity, and suitability for graduate-level examinations similar to those written by faculty. In the context of English education, Lee et al. \cite{lee2024applying} demonstrated that prompt-engineered LLMs could generate questions with statistically significant validity, highlighting the practical potential of few-shot and role-based prompting strategies.

Despite these promising results, prompt-based single-pass generation exhibits important limitations. Question quality is often highly sensitive to prompt formulation \cite{reynolds2021prompt, wei2022chain}, and reliably eliciting higher-order cognitive demands—such as inference or synthesis—remains challenging without additional structural constraints \cite{scaria2024automated, moore2023developing}. Moreover, single-pass prompting provides limited mechanisms for systematic verification, revision, or enforcement of constraints on distractor quality and stylistic consistency. These limitations motivate system-level approaches that integrate structured control and iterative refinement, moving beyond the treatment of LLMs as monolithic generators.

\subsection{Hybrid and Multi-agent Approaches}
As transformer and LLM-based methods have matured, research has increasingly shifted toward hybrid frameworks that combine learning-based models with rule-based components, external knowledge sources, or structured control mechanisms. Rather than treating language models as fully self-sufficient generators, these approaches augment LLMs with additional structure, exemplars, or externally computed constraints to improve reliability, interpretability, and pedagogical alignment.

One prominent line of work integrates external knowledge representations with language models to support more complex forms of question generation. For example, Li et al. \cite{li2023multi} combined language modeling with knowledge graph construction to enable multi-hop question generation, allowing questions to be generated through reasoning across multiple related facts rather than single-sentence transformations. Similarly, Hang et al. \cite{hang2024mcqgen} proposed an LLM-driven MCQ generation framework that incorporates retrieval-augmented generation (RAG), enabling the system to draw on an external knowledge base to better align generated questions with instructional objectives and learner needs. 

Another stream of hybrid research has focused on distractor generation, a long-standing challenge in MCQ authoring. Bitew et al. \cite{bitew2023distractor} introduced a predictive prompting strategy in which relevant items are retrieved from an existing question bank and used as in-context demonstrations to guide LLMs toward producing more plausible distractors. Their results showed improvements over zero-shot and static few-shot prompting baselines in expert evaluations. Similarly, Yu et al. \cite{yu2024enhancing} proposed a retrieval-augmented pretraining approach that integrates knowledge graphs and task-specific pretraining to enhance distractor generation. By retrieving relevant passages and knowledge triplets to construct pseudo-questions, their method improved both distractor quality and semantic relevance. 

In parallel with these hybrid enhancements, agentic and multi-agent systems have emerged as a practical architecture for orchestrating LLMs on complex, multi-step tasks \cite{christie2025agentic, lu2025octotools}. In such systems, specialized agents—such as planners, generators, reviewers, and external tools—are coordinated to decompose complex workflows into tractable subtasks while combining deterministic rule-based processing with stochastic LLM outputs. This paradigm has gained traction as a means of improving controllability, robustness, and scalability in LLM-driven applications.

Recent studies demonstrate the promise of multi-agent pipelines for automated MCQ generation. Bhowmick et al. \cite{bhowmick2023automating}, for instance, proposed a modular system with separate components for question generation, correct answer prediction, and distractor formulation, implemented using transformer-based models such as T5 and GPT-3. Pawar et al. \cite{pawar2024automated} introduced a LangChain-based multi-step framework that integrates text extraction, MCQ generation, complexity evaluation, review generation, and result presentation to produce instructor-usable quizzes with minimal human revision. Mucciaccia et al. \cite{mucciaccia2025automatic} presented an LLM-based agentic system in which preprocessing, glossary generation, preliminary question generation, and question review are handled by distinct modules, reporting that the system produced high-quality questions suitable for academic use. Collectively, these systems highlight the advantages of modular design, explicit role assignment, and iterative evaluation in MCQ generation workflows.

Despite these advances, persistent challenges remain. Empirical studies indicate that while AI-generated MCQs are often grammatically correct and fluent, they frequently overemphasize trivial details, lack conciseness, or include distractors that are obviously incorrect, stylistically inconsistent, or insufficiently parallel to the correct answer \cite{gorgun2024exploring, ling2024automatic, lee2024math}. Most importantly, existing hybrid and agentic frameworks have yet to demonstrate reliable control over the cognitive complexity of generated questions. Empirical evaluations suggest that alignment with higher-order cognitive levels (e.g., inferential or synthesis questions) remains inconsistent without carefully designed control mechanisms \cite{scaria2024automated}. These limitations highlight the need for frameworks that integrate agentic decomposition with hybrid control mechanisms that combine LLM-based generation and deterministic rule-based constraints, enabling explicit cognitive targeting and iterative refinement. This motivates approaches such as ReQUESTA that emphasize control over how questions are generated, not merely what is generated.

\section{The ReQUESTA Framework}
\subsection{System Overview and Workflow}

\begin{figure}
\includegraphics[width=\textwidth]{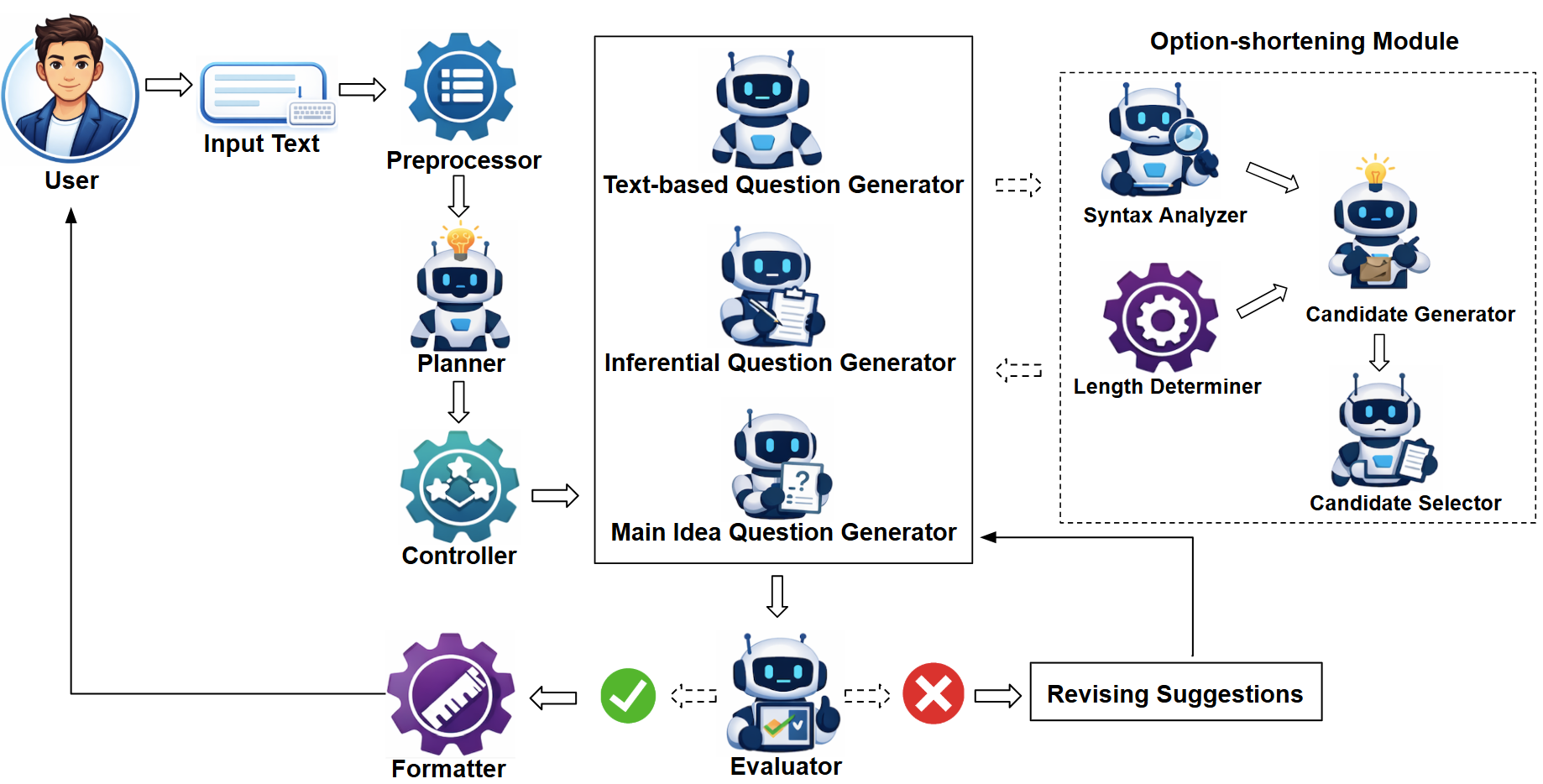}
\caption{Overview of the ReQUESTA hybrid agentic workflow. The system implements a multi-stage, modular pipeline that combines LLM-powered agents with rule-based components to support controlled, reliable MCQ generation. LLM-powered agents are depicted as robots, while rule-based agents are shown as gears.} \label{fig1}
\end{figure}

ReQUESTA, illustrated in Fig.~\ref{fig1}, is a hybrid agentic framework for controllable MCQ generation, designed to combine the expressive power of LLMs with the reliability of deterministic, rule-based processing. From a systems perspective, ReQUESTA treats MCQ generation as a constrained, multi-stage generation problem rather than a single text-to-output transformation. The system accepts an input text together with configuration parameters (e.g., the number and types of questions to be generated) and produces a finalized set of MCQs that satisfy structural constraints and specified cognitive demands. Internally, this process is implemented through a coordinated sequence of LLM-powered agents, responsible for open-ended language generation and interpretation, and rule-based agents, responsible for deterministic coordination, constraint enforcement, and post-processing. This explicit separation allows stochastic generation to be paired with deterministic control, reducing variance and improving reliability in the generated outputs.

The workflow begins with a Preprocessor, a rule-based agent that segments the input text into coherent units using sentence-level parsing. This deterministic preprocessing step ensures that downstream agents operate on text segments that are contextually meaningful and of manageable length for LLM processing, while preserving local coherence within each segment.

The segmented text is then passed to the Planner, an LLM-powered agent that performs high-level analytical and organizational functions. Guided by structured, stepwise prompting, the Planner summarizes each segment, identifies key concepts and implicit inferences, and constructs a question-generation plan. This plan specifies which segments and extracted elements will be used to generate different types of questions (text-based, inferential, or main idea) in accordance with user requests. Rather than generating questions directly, the Planner produces an intermediate representation that encodes both the intended cognitive focus and the allocation of generation tasks.

Next, the Controller, a rule-based coordination agent implemented with custom Python scripts, interprets the Planner’s output and orchestrates task execution. The Controller assigns the relevant text segments, along with the associated key concepts and inferences, to the appropriate Question Generator agents. By separating planning (probabilistic) from execution control (deterministic), ReQUESTA ensures that downstream generation remains predictable and auditable.

Question generation is carried out by three specialized Question Generator agents, each implemented as an LLM-powered agent with distinct prompting constraints. The Text-based Question Generator produces questions targeting explicit concepts and factual information stated in the text. The Inferential Question Generator focuses on relationships and reasoning that require integrating information beyond surface-level recall. The Main Idea Question Generator generates questions that assess understanding of overarching themes or central arguments. Together, these agents operate on scoped inputs defined by the Planner and Controller, reducing prompt ambiguity and cross-task interference.

All generated questions are subsequently processed by an Evaluator, an LLM-powered agent that assesses item quality using a predefined evaluation checklist. The criteria include clarity of the question stem, alignment between the stem and the correct answer, and the plausibility of distractors. Questions that satisfy these criteria are approved for output, whereas those that do not are automatically returned to the relevant Question Generator along with targeted revision suggestions. This mechanism embeds iterative refinement directly into the generation pipeline, rather than relying on manual post-hoc review.

Following evaluation, a Formatter, implemented as a rule-based agent, applies post-processing operations. The Formatter standardizes question presentation by shuffling answer options, enforcing consistent labeling, and applying uniform formatting conventions across all items.

In addition to the core pipeline, ReQUESTA incorporates an Option-Shortening Module to address length imbalances among answer options—a common failure mode in LLM-generated MCQs. This module detects options that are substantially longer than others and revises them while preserving their original meaning. It consists of four collaborating agents: a Syntax Analyzer that leverages an LLM to identify shared syntactic patterns across options; a Length Determiner, a rule-based agent that establishes an appropriate length range based on the remaining options; an LLM-powered Candidate Generator that produces concise alternative phrasings; and a Candidate Selector that evaluates these candidates using an LLM informed by rule-based metrics such as word count and semantic similarity. The Option-Shortening Module operates as a post-processing step prior to final evaluation and formatting.

Taken together, these components form a controlled generation pipeline in which uncertainty is localized to specific agents, constraints are enforced deterministically, and intermediate representations mediate complex task execution. Subsequent sections elaborate on the design rationale underlying this architecture and the prompt engineering strategies used to implement individual agents.

\subsection{Design Rationale and Agentic Principles}
ReQUESTA’s architecture is guided by AI engineering principles aimed at controlling stochastic generation in complex, multi-step tasks. Rather than treating MCQ generation as a single, monolithic text-to-output task, the framework decomposes the process into a sequence of well-defined subtasks, each handled by a specialized agent, allowing probabilistic and deterministic components to complement one another.

\subsubsection{Decomposition and Control}
MCQ authoring involves heterogeneous subtasks, including text analysis, identification of assessable content, formulation of question stems, construction of plausible distractors, and post-processing to ensure consistency and usability. In ReQUESTA, these subtasks are explicitly designated to specific agents responsible for text preprocessing, key information identification and extraction, question generation, evaluation, and post-processing. This decomposition enables fine-grained control over each stage of the pipeline and allows individual components to be designed, tested, and adjusted independently.

By separating planning from execution and isolating cognitively distinct generation tasks, the framework reduces the reliance on implicit reasoning within a single LLM prompt. Instead, cognitive intent is encoded in intermediate representations that guide downstream agents. This structure supports more predictable and controllable generation behavior, particularly when targeting specific cognitive focuses such as recall, inference, or synthesis.

\subsubsection{Iterative Refinement}
A second guiding principle of ReQUESTA is embedded iterative refinement. LLM-generated questions may occasionally fail to meet basic structural or pedagogical requirements, such as producing ambiguous stems or implausible distractors. To address this, the framework incorporates automatic failure detection and retry mechanisms within the generation process. For example, when an initial generation attempt does not yield a complete MCQ with a valid stem and four options, the system initiates a new generation cycle without requiring user intervention.

Beyond structural completeness, iterative interaction between the Evaluator and the Question Generator agents further refines question quality. Generated items are evaluated against a predefined checklist, and those that do not meet the criteria are returned to the corresponding generator along with targeted revision feedback. This feedback loop continues until the item satisfies the quality thresholds or reaches a termination condition. Through this mechanism, refinement is embedded directly into the workflow rather than deferred to manual review.

\subsubsection{Parallel Generation} 
ReQUESTA also leverages parallel generation to improve efficiency. To illustrate, the three Question Generator agents (text-based, inferential, and main idea) operate concurrently once the generation plan has been established. Because each generator is assigned a distinct cognitive role and operates on scoped inputs, questions can be produced in parallel without interference. This design reduces overall latency and enables the system to scale to larger input texts or higher requested numbers of questions while maintaining consistent cognitive targeting.

\subsubsection{Modular Extensibility} 
Finally, ReQUESTA is designed as a modular and extensible system in which individual components can be replaced, modified, or augmented without restructuring the entire pipeline. Each agent operates as an independent module with clearly defined inputs and outputs, facilitating experimentation with alternative implementations or additional functionality. For example, the Option-Shortening Module was incorporated after the core architecture had been completed to address stylistic imbalances among answer options. Its integration required no changes to the upstream planning or generation components, illustrating the flexibility of the overall design.

This modularity supports future extensions, such as introducing new cognitive categories, integrating alternative evaluation metrics, or replacing LLM components as models evolve. Together, these design principles (i.e., decomposition and control, iterative refinement, parallel generation, and modular extensibility) define ReQUESTA’s agentic approach and distinguish it from end-to-end prompt-based systems for automated question generation.

\subsection{Prompt Engineering in ReQUESTA}
In the current ReQUESTA architecture, state-of-the-art prompt engineering techniques are employed to augment the capacity of LLM-powered agents. Rather than treating prompts as static instructions, ReQUESTA dynamically tailors prompts to shape how LLMs interpret tasks, allocate attention, and structure their reasoning processes. Specifically, prompts are constructed and augmented at runtime using information produced by other agents or external computational functions, allowing each agent to operate with contextually relevant, task-specific signals. This dynamic prompt augmentation enables ReQUESTA to achieve more controlled and context-sensitive generation behavior.

Overall, four primary prompt engineering strategies are used across the framework: persona-based prompting, chain-of-thought (CoT) prompting, few-shot prompting, and self-critique.

\subsubsection{Persona-based Prompting}
Persona-based prompting has been shown to influence model behavior and output quality by providing contextual grounding for task interpretation \cite{reynolds2021prompt, schmidt2023cataloging, wang2023unleashing}. In ReQUESTA, personas are used to align model outputs with expectations drawn from human assessment practices. Each LLM-powered agent in ReQUESTA is assigned a specific role, or persona, that defines its functional identity and scope of expertise. 

For example, all Question Generator agents share a common role description: “\textit{You are an experienced college instructor and an expert in writing multiple-choice questions to assess students’ understanding of academic texts (e.g., textbook chapters, academic articles).}” This role anchors generation behavior in instructional and assessment norms, encouraging the production of questions that resemble those authored by subject-matter experts rather than generic or surface-level prompts.

Other agents are assigned more specialized roles aligned with their functional objectives. For example, the Syntax Analyzer in the Option-Shortening Module adopts the role of “\textit{a linguist and expert in syntactic analysis of educational content.}” This role primes the model to focus on structural and grammatical patterns rather than semantic interpretation, supporting consistent and linguistically informed option revision. By differentiating roles across agents, ReQUESTA constrains LLM behavior in ways that reflect the division of labor within the overall framework.

\subsubsection{CoT Prompting} 
CoT prompting is employed in ReQUESTA to elicit explicit, step-by-step reasoning from LLM-powered agents tasked with analytical, planning, and evaluative functions \cite{wei2022chain}. In this framework, CoT prompting serves two complementary purposes: it specifies an ordered sequence of task execution steps and requires agents to report their intermediate reasoning as part of the output. This combination ensures that complex subtasks are executed in a structured manner while making the underlying reasoning process explicit and inspectable.

To illustrate, the Planner is instructed to carry out the question-planning process through a clearly defined sequence of steps: (1) summarizing the input text, (2) identifying key facts and/or inferences, (3) organizing these elements according to their importance for understanding the text, (4) selecting a subset of facts or inferences based on user-specified requirements, and (5) producing the final question-generation plan in a structured JSON format. In addition to completing these steps, the Planner is prompted to explicitly report its reasoning at each stage, thereby externalizing the decision process that governs content selection and cognitive targeting. By requiring agents to both follow prescribed steps and articulate their reasoning, CoT prompting reduces ambiguity in ReQUESTA task execution and enforces procedural structure while simultaneously enabling interpretable reasoning within LLM-driven components.

\subsubsection{Few-shot Prompting}  
Few-shot prompting is used in ReQUESTA to calibrate LLM outputs by providing illustrative examples of desired and undesired responses \cite{brown2020language}. This strategy supports inductive learning of task-specific constraints that are difficult to encode through rules alone, particularly when stylistic or semantic nuances are involved.

Within ReQUESTA, few-shot prompting is applied selectively rather than uniformly across all agents. For example, in the Candidate Generator of the Option-Shortening Module, the LLM is provided with examples of both effective and ineffective shortened answer options. These examples demonstrate how excessive length can be reduced while preserving semantic equivalence and grammatical integrity. By contrasting positive and negative examples, the agent is guided toward producing revisions that balance conciseness with meaning preservation.

\subsubsection{Self-critique Prompting}   
In addition to externally applied evaluation, ReQUESTA incorporates self-critique prompting to enable LLM-powered agents to reflect on and assess their own outputs. Prior work has shown that self-critique can improve generation quality by encouraging models to engage in metacognitive monitoring and internal error detection \cite{huang2023large}. In ReQUESTA, self-critique functions as an internal quality-control mechanism that precedes and complements external evaluation.

For instance, each Question Generator in ReQUESTA is prompted with a set of diagnostic questions immediately after generating an MCQ. These prompts are designed to surface common weaknesses in multiple-choice items, such as ambiguity in the question stem, misalignment between the stem and the correct answer, or implausible distractors. Example self-critique questions include, \textit{Is the question stem clear and unambiguous? Is the correct answer clearly the best option? Are the distractors plausible and relevant?} Agents are instructed to review their outputs in light of these criteria and revise the question if deficiencies are identified. 

By embedding self-critique directly within the generation process, ReQUESTA reduces the likelihood that low-quality items propagate downstream prior to the Evaluator agent described in Section 3.1. Together, self-critique prompting and evaluator-based feedback form a layered quality assurance mechanism within the overall agentic workflow.

\section{Experimental Methodology}

\subsection{Participants}
Participants were recruited through Prolific, a widely used crowdsourcing platform for academic research. Eligibility criteria required that participants (a) were at least 18 years old, (b) currently resided in the United States, and (c) had a Prolific approval rate of 98\% or higher. Participants received \$6.00 in compensation for completing the study, and the mean completion time for the study was approximately 25 minutes.

To ensure data quality, multiple exclusion criteria were applied prior to analysis. Participants were removed if they failed embedded attention checks, exhibited inattentive response patterns, or displayed implausibly short completion times indicative of clicking-through behavior. Specifically, participants who spent less than 10 seconds on at least one passage block were excluded. In addition, automated or non-human responses were detected and removed using Google’s reCAPTCHA system.

After applying these criteria, the final analytic sample consisted of 572 participants with valid data. Participants ranged in age from 18 to 77 years ($M = 41.01$, $SD = 12.82$). The sample included 305 male participants (53.32\%), 255 female participants (44.58\%), and 12 participants identifying as another gender (2.1\%). With respect to educational attainment, 321 participants (56.12\%) reported holding a bachelor’s degree or higher. The majority of participants (96.5\%, $n = 552$) reported English as their native language.

\subsection{Text Selection for Question Generation}
Twenty academic expository passages were selected from OpenStax textbooks in Sociology, Lifespan Development, History, and Anthropology. Each passage was approximately 400 words in length. Passages were matched across domains based on word count and readability indices (e.g., Flesch--Kincaid Grade Level) to ensure comparable linguistic complexity.

\subsection{Question Generation Procedures}
For each passage, two parallel sets of MCQs were generated. One set was generated using ReQUESTA while the second set was generated using a GPT-5 zero-shot baseline. The baseline employed a single instruction prompt designed to approximate a typical, minimally engineered prompt that a non-expert user might provide in practice. This prompt specified only the definitions of the three question types (text-based, inferential, and main idea), using the same definitions applied within ReQUESTA, but did not include worked examples, multi-step instructions, or explicit guidelines for MCQ authoring (e.g., distractor construction or stylistic constraints). 

Both ReQUESTA and the baseline system operated on the same source passages and produced MCQs with an identical target composition: two text-based questions, two inferential questions, and one main idea question per passage. ReQUESTA used GPT-5 as the underlying language model for its LLM-powered agents; however, unlike the baseline condition, question generation in ReQUESTA proceeded through a multi-step workflow that included intermediate planning representations, specialized question generators, automated evaluation, and iterative refinement. The baseline GPT-5 condition, by contrast, reflected a typical single-model usage scenario in which all questions were generated in a single pass without external control mechanisms or revision loops. This design provides a realistic single-pass baseline and enables comparison between end-to-end prompting and a hybrid, agentic workflow built around planning, coordination, and revision, while holding the underlying language model constant.

Across 20 passages, this procedure yielded a total pool of 200 MCQs (20 passages × 5 questions × 2 generation methods), balanced across four academic domains. Representative examples of MCQs generated by ReQUESTA and the GPT-5 baseline, along with their source passages, are provided in the Supplementary Materials.

\subsection{Reading Comprehension Task and Experimental Design}
Participants were invited to a custom web-based application\footnote{The source code for the web application used in this study is publicly available at \url{https://github.com/terryyutian/requesta_evaluation/tree/main}.} to complete a reading comprehension task consisting of three passage blocks. In each block, participants read one passage and answered five MCQs generated by a single question-generation method (either ReQUESTA or GPT-5). Passage assignment and question source were randomized with one constraint: each participant was exposed to both generation methods. Specifically, participants received either two passages paired with ReQUESTA and one paired with GPT-5, or the reverse. This design enabled within-participant comparison of question source while limiting task duration and fatigue.

Within each passage block, the cognitive composition of questions was held constant (two text-based, two inferential, one main idea). No time limits were imposed. Participants were allowed to revisit the passage while answering questions and could revise their responses prior to submission.

Overall, the experiment employed a within-participant, passage-level comparison of question source, with question source manipulated at the passage block level and question type held constant within each block. Because participants responded to multiple items and each item was answered by multiple participants, responses were cross-classified and clustered within participants and items. This dependency structure was explicitly modeled in subsequent analyses using mixed-effects methods.

\subsection{Vocabulary Assessment}
To control for individual differences in lexical proficiency, participants completed an adapted version of the Lexical Test for Advanced Learners of English (LexTALE; \cite{lemhofer2012introducing}) before the reading comprehension task. The task consisted of 60 items (40 real English words, 20 pseudowords), presented sequentially on the screen. Participants had one minute to identify each item as a real word or non-word by selecting \textsc{yes} or \textsc{no}. Vocabulary scores were calculated as the total number of correct responses within the time limit.

\subsection{Psychometric Analysis}

\subsubsection{Classical Test Theory (CTT) Analysis}

Item-level psychometric properties were examined using indices from CTT as descriptive diagnostics of question quality. Given the incomplete-block design of the study in which each participant responded to a subset of items rather than a single, fixed test form, CTT indices were computed at the item level using responses from all participants who encountered each item, rather than treating the data as arising from a single coherent test.

\paragraph{Item Difficulty ($p$).} Item difficulty was operationalized as the proportion of participants who answered the item correctly among those who responded to that item. Values ranged from 0 to 1, with higher values indicating easier items.

\paragraph{Item Discrimination ($D$).} Item discrimination was assessed using the discrimination index, defined as the difference in the proportion of correct responses between participants in the upper and lower quartiles of the total reading comprehension score distribution. Total scores were calculated as each participant’s sum of correct responses across the items they completed. Although participants encountered different item sets, this index provides a useful approximation of how well each item differentiates relatively higher- and lower-performing readers within the sampled population.

\paragraph{Point-Biserial Correlation ($r_{pb}$).} Point-biserial correlations were computed between each item’s correctness (0/1) and participants’ total test scores across the items they completed, with the target item removed from the total score to avoid part–whole inflation. The point-biserial index estimates the extent to which performance on a given item aligns with overall reading comprehension performance within the sampled design.

\subsubsection{Mixed-Effects Logistic Regression Modeling}

Item-level accuracy (0 = incorrect, 1 = correct) was modeled using mixed-effects logistic regression to account for the crossed hierarchical structure of the data, with multiple participants responding to multiple items. All analyses were conducted in R \cite{r2016r} using the \texttt{lme4} package \cite{bates2015fitting}. Two random intercepts were included: Participants (representing individual differences in reading ability and test-taking performance) and Items (representing systematic variability in item difficulty).

The primary goal of the analysis was to evaluate whether item accuracy differed as a function of question source (ReQUESTA vs.\ GPT-5), question type (text-based, inferential, main idea), and participants’ vocabulary knowledge (centered LexTALE scores), as well as whether these predictors interacted. Model building proceeded in a hierarchical, stepwise manner. First, a baseline random-intercepts model (M0) with no fixed predictors was estimated. Second, a main-effects model (M1) was fit that included fixed effects for question source, question type, and centered vocabulary ability. Third, a model including all possible two-way interactions among these predictors (M2) was evaluated. Finally, a full model with a three-way interaction between source, question type, and vocabulary ability (M3) was estimated.

During model development, we also explored alternative random-effects structures that included random slopes for question source and question type by participant, as well as by item, to account for potential individual differences in sensitivity to question source or cognitive focus. However, these models either failed to converge, resulted in singular fits, or did not yield substantive improvements in model fit. In light of these considerations and in the interest of parsimony and model stability, we retained a random-intercepts structure for the final analyses.

Models were compared using likelihood ratio tests (LRTs) and changes in Akaike Information Criterion (AIC). Following standard recommendations for mixed-effects modeling in educational measurement, the simplest model that provided a statistically and substantively meaningful improvement in fit was selected as the final model. Interaction terms were retained only if they significantly improved model fit and contributed unique explanatory value. All models were estimated using the \texttt{nloptwrap} optimizer with increased iteration limits to ensure stable convergence.

\subsection{Expert Evaluation}
\subsubsection{Evaluation Rubric}
To assess the quality of MCQs generated by ReQUESTA and GPT-5, a rubric was designed to evaluate key aspects of high-quality MCQs: answer correctness, distractor incorrectness, topic relevance, writing clarity, distractor linguistic features, distractor semantic plausibility, and distractor semantic uniqueness (see Appendix A for the full rubric). \textit{Answer correctness} was defined as whether the “correct answer is clearly correct to readers who have comprehended the text,” and \textit{distractor incorrectness} as whether “distractors are clearly incorrect to readers who have comprehended the text.” Because GenAI-generated MCQs rarely include inaccurate answers or accurate distractors, these two criteria were scored dichotomously (0 or 1). 

The remaining criteria are more challenging for LLMs and were therefore scored on a 4-point scale (1 to 4). This afforded more nuanced judgments than a binary or 3-point scale, and reduced the tendency for raters to default to a middle option. \textit{Topic relevance} referred to the extent to which an MCQ addressed a central concept in the text. \textit{Writing clarity} captured the extent to which the question stem and response options were concise and unambiguous. \textit{Distractor linguistic features} assessed the similarity of distractors in attributes such as length, number of clauses, and overall structure. \textit{Distractor semantic plausibility} reflected the extent to which distractors were reasonable to readers who failed to fully comprehend the text (e.g., topic alignment with the correct answer, avoiding absolute terms). \textit{Distractor semantic uniqueness} referred to the extent to which each distractor conveyed a unique meaning, as distractors with similar meanings are easy for readers to eliminate. 

\subsubsection{Evaluation Procedure}
Two expert raters initially calibrated using a set of 100 practice MCQs. Over three rounds of scoring, raters iteratively refined the rubric to ensure that it both targeted key flaws in generated questions and maximized reliability between raters. Then both raters scored the current study’s set of 200 questions. Question order was randomized, and question source was blinded. Reliability was calculated after every 50 scored questions. Following each calculation, raters met to discuss disagreements, occasionally consulting a third rater to resolve disagreements, and then both raters separately revised their scores. The final Cohen’s weighted kappa (distractor incorrectness: $\kappa$ = 1; overall relevance: $\kappa$ = .80; writing clarity: $\kappa$ = .88; distractor linguistic features: $\kappa$ = .85;	distractor semantic plausibility: $\kappa$ = .80;	distractor semantic uniqueness: $\kappa$ = .91) was calculated for an estimate of the interrater reliability (Cohen, 1960). All questions received a score of 1 for answer correctness. Remaining disagreements were resolved by adopting the lower score to ensure conservative quality standards, given the overall high quality of the MCQs.

\section{Results}

\subsection{Results from Psychometric Analyses}

\subsubsection{CTT Results}

Descriptive statistics for item difficulty, item discrimination, and point-biserial correlations are presented in Table~\ref{tab:ctt}. As shown, ReQUESTA items had a lower mean difficulty index ($p = .786$, $SD = .133$) than GPT-5 items ($p = .902$, $SD = .112$), indicating that participants found ReQUESTA-generated questions more challenging. 

In terms of discrimination, ReQUESTA items showed substantially higher discrimination values ($M = .39$, $SD = .196$) compared to GPT-5 items ($M = .17$, $SD = .131$), suggesting that ReQUESTA items better differentiated higher- and lower-performing participants. 

Point-biserial correlations showed a similar pattern: ReQUESTA items ($M = .323$, $SD = .168$) displayed stronger associations with total test scores than GPT-5 items ($M = .275$, $SD = .202$), indicating higher alignment with underlying reading comprehension ability.

\begin{table}[t]
\caption{Descriptive statistics for CTT item metrics.}
\label{tab:ctt}
\centering
\begin{tabular}{|l|cc|cc|cc|}
\hline
\textbf{Source} & \multicolumn{2}{c|}{\textbf{Difficulty ($p$)}} & 
\multicolumn{2}{c|}{\textbf{Discrimination ($D$)}} & 
\multicolumn{2}{c|}{\textbf{Point-biserial ($r_{pb}$)}} \\
 & M & SD & M & SD & M & SD \\
\hline
GPT-5 & 0.902 & 0.112 & 0.17 & 0.131 & 0.275 & 0.202 \\
ReQUESTA & 0.786 & 0.133 & 0.39 & 0.196 & 0.323 & 0.168 \\
\hline
\end{tabular}
\end{table}

\subsubsection{Mixed-Effects Logistic Regression}

The main-effects model (M1), which included fixed effects for question source, question type, and centered vocabulary ability, provided a substantial improvement in fit over the baseline random-intercepts model (M0), $\Delta\chi^{2}(4) = 77.23$, $p < .001$, and showed a markedly lower AIC. Adding two-way interactions (M2) did not significantly improve model fit relative to M1, $\Delta\chi^{2}(5) = 9.57$, $p = .088$, and resulted in a slightly higher AIC. Similarly, the full three-way interaction model (M3) did not improve model fit over the main-effects model, $\Delta\chi^{2}(7) = 13.20$, $p = .067$, and again yielded a higher AIC. Accordingly, the main-effects model (M1) was selected as the final, best-fitting model.

\begin{table}[t]
\caption{Fixed effects from mixed-effects logistic regression predicting correct responses ($N = 8{,}440$).}
\label{tab:mixedeffects}
\centering
\begin{tabular}{|l|c|c|c|c|}
\hline
\textbf{Fixed Effect} & \textbf{Coefficient} & \textbf{SE} & \textbf{z} & \textbf{p} \\
\hline
Intercept & 2.834 & 0.157 & 18.00 & <.001 \\
Source: ReQUESTA (vs.\ GPT-5) & -1.277 & 0.166 & -7.687 & <.001 \\
Question Type: main idea (vs.\ inferential) & 0.323 & 0.190 & 1.701 & .090 \\
Question Type: text-based (vs.\ inferential) & 0.385 & 0.160 & 2.400 & .017 \\
Vocabulary & 0.025 & 0.006 & 4.414 & <.001 \\
\hline
\end{tabular}

\vspace{2mm}
\noindent\textit{Note.} Marginal $R^{2} = .085$, conditional $R^{2} = .448$.

\end{table}

\begin{figure}
\includegraphics[width=0.85\textwidth]{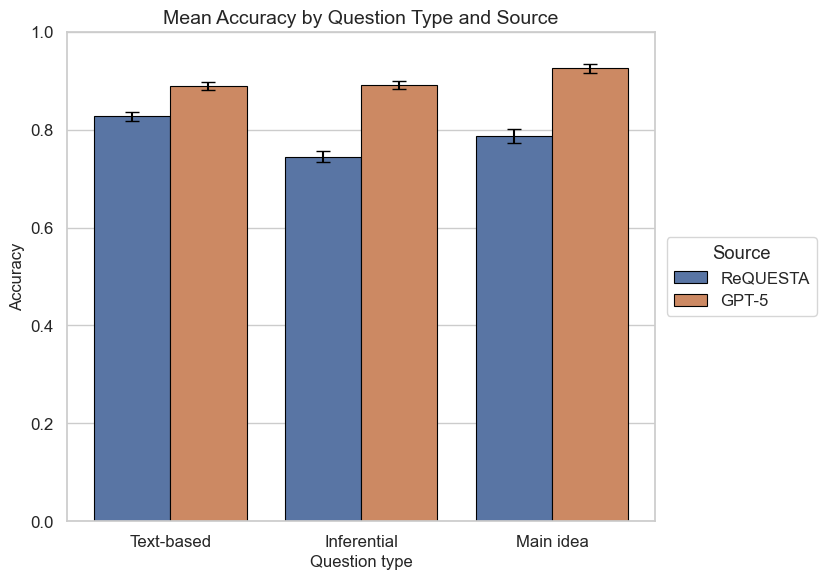}
\caption{Mean accuracy for MCQs by question type (text-based, inferential, main idea) and question source (ReQUESTA vs. GPT-5). Error bars represent ±1 standard error. ReQUESTA-generated items show consistently lower accuracy across all question types, indicating higher difficulty compared to GPT-5-generated items.} \label{fig2}
\end{figure}

Table~\ref{tab:mixedeffects} presents the fixed effects from the final mixed-effects logistic regression model predicting the likelihood of a correct response. Question source was a strong and significant predictor of accuracy. Items generated by ReQUESTA showed a markedly lower probability of a correct response ($\beta = -1.277$, $p < .001$), suggesting that they were substantially more difficult than those produced by GPT-5. As illustrated in Fig.~\ref{fig2}, this pattern held consistently across all question types.

Question type also contributed to performance differences. Compared with inferential questions, text-based questions were significantly easier ($\beta = 0.385$, $p = .017$). Participants’ vocabulary knowledge significantly predicted accuracy across all items ($\beta = 0.025$, $p < .001$), indicating that individuals with higher lexical proficiency were more likely to answer items correctly regardless of question type or source.

Overall model fit indices showed that the fixed effects explained a modest proportion of variance in accuracy (marginal $R^{2} = .085$), whereas the inclusion of random intercepts for participants and items accounted for substantial additional variability (conditional $R^{2} = .448$). This pattern indicates that, although source, question type, and vocabulary ability reliably influenced performance, considerable variability remained attributable to individual differences and inherent differences in item difficulty.

\subsection{Expert Evaluation Results}
Answer correctness (\textit{M} = 1.00, \textit{SD} = 0.00), distractor incorrectness (\textit{M} = 0.99, \textit{SD} = 0.10), topic relevance (\textit{M} = 3.10, \textit{SD} = 0.94), writing clarity (\textit{M} = 3.20, \textit{SD} = 0.72), distractor linguistic features (\textit{M} = 2.23, \textit{SD} = 1.15), distractor semantic plausibility (\textit{M} = 2.91, \textit{SD} = 0.95), and distractor semantic uniqueness (\textit{M} = 3.71, \textit{SD} = 0.61) were scored across 100 GPT-5 generated items and 100 ReQUESTA generated items (see Table~\ref{tab:expertdescriptives}). A two-way multivariate ANOVA was conducted to examine the effects of question source (ReQUESTA vs. GPT-5) and question type (text-based, inferential, main idea) on question quality scores (topic relevance, writing clarity, distractor linguistic features, distractor semantic plausibility, and distractor semantic uniqueness). Answer correctness and distractor incorrectness rarely received scores other than 1 and were not included in the multivariate ANOVA. Prior to conducting the analysis, multivariate outliers were tested for using Mahalanobis distance; three items were removed from analysis. Analyses were conducted using SPSS (Version 29).

\begin{table}[t]
\caption{Descriptive statistics for scoring category, separated by question source and question type. \textit{N} = 200.}
\label{tab:expertdescriptives}
\centering
\begin{tabular}{|l|l|c|c|c|}
\hline
\textbf{Scoring Category} & \textbf{Question Source} &
\textbf{Text-Based} & \textbf{Inferential} & \textbf{Main Idea} \\
 &  & \textbf{\textit{M} (\textit{SD})} & \textbf{\textit{M} (\textit{SD})} & \textbf{\textit{M} (\textit{SD})} \\
\hline
\multirow[t]{2}{*}{Answer Correctness} 
& ReQUESTA & 1.00 (0.00) & 1.00 (0.00) & 1.00 (0.00) \\
& GPT-5 & 1.00 (0.00) & 1.00 (0.00) & 1.00 (0.00) \\
\hline
\multirow[t]{2}{*}{Distractor Incorrectness} 
& ReQUESTA & 1.00 (0.02) & 1.00 (0.02) & 1.00 (0.02) \\
& GPT-5 & 0.98 (0.02) & 0.98 (0.02) & 1.00 (0.02) \\
\hline
\multirow[t]{2}{*}{Topic Relevance} 
& ReQUESTA & 2.73 (0.12) & 3.40 (.12) & 3.95 (0.17) \\
& GPT-5 & 2.28 (0.12) & 3.13 (0.12) & 3.95 (0.17) \\
\hline
\multirow[t]{2}{*}{Writing Clarity} 
& ReQUESTA & 3.40 (0.11) & 2.70 (0.11) & 2.95 (0.15) \\
& GPT-5 & 3.38 (0.11) & 3.28 (0.11) & 3.5 (0.15) \\
\hline
\multirow[t]{2}{*}{Distractor Linguistic Features} 
& ReQUESTA & 2.85 (0.16) & 2.43 (0.16) & 1.5 (0.23) \\
& GPT-5 & 2.58 (0.16) & 1.88 (0.16) & 1.35 (0.23) \\
\hline
\multirow[t]{2}{*}{Distractor Semantic Plausibility} 
& ReQUESTA & 3.13 (0.15) & 3.13 (0.15) & 3.30 (0.21) \\
& GPT-5 & 2.85 (0.15) & 2.60 (0.15) & 2.40 (0.21) \\
\hline
\multirow[t]{2}{*}{Distractor Semantic Uniqueness} 
& ReQUESTA & 3.7 (0.09) & 3.63 (0.09) & 4.00 (0.13) \\
& GPT-5 & 3.83 (0.09) & 3.48 (0.09) & 3.9 (0.13) \\
\hline
\end{tabular}
\end{table}

The multivariate main effect of question source was significant, Pillai's Trace = .196, F(5, 187) = 9.144, \textit{p} < .001, partial $\eta^2$ = .196 (see Table~\ref{tab:manova}). Overall, ReQUESTA-generated MCQs received higher scores for topic relevance, distractor linguistic features, and distractor semantic plausibility (all \textit{p}s < .05). In contrast, GPT-5 MCQs received higher scores for writing clarity, \textit{p} < .001. No significant differences were observed between ReQUESTA and GPT-5 for answer correctness, distractor incorrectness, and distractor semantic uniqueness.

\begin{table}[t]
\caption{Post-hoc analyses of the effect of question source on question quality scores. *\textit{p} < .05. ***\textit{p} < .001}
\label{tab:manova}
\centering
\begin{tabular}{|l|c|c|c|c|c|c|}
\hline
\multirow{2}{*}{\textbf{Scoring Category}} 
& \multicolumn{2}{c|}{\textbf{ReQUESTA}} 
& \multicolumn{2}{c|}{\textbf{GPT-5}} 
& \multirow{2}{*}{\textbf{\textit{F}(1,191)}} 
& \multirow{2}{*}{\textbf{$\eta^2$}} \\
\cline{2-5}
& \textbf{\textit{M}} & \textbf{\textit{SE}} 
& \textbf{\textit{M}} & \textbf{\textit{SE}} 
&  &  \\
\hline
Topic Relevance & 3.36 & 0.08 & 3.12 & 0.08 & 4.54 & .02* \\
Writing Clarity & 3.02 & 0.07 & 3.39 & 0.07 & 13.27 & .07*** \\
Distractor Linguistic Features & 2.28 & 0.11 & 1.92 & 0.11 & 5.59 & .03* \\
Distractor Semantic Plausibility & 3.17 & 0.10 & 2.63 & 0.10 & 15.57 & .08*** \\
Distractor Semantic Uniqueness & 3.82 & 0.05 & 3.75 & 0.05 & 0.80 & .00 \\
\hline
\end{tabular}
\end{table}

Examination of the Source × Question Type interaction revealed that question quality scores differed by question source for each question type, Pillai's Trace = .098, F(10, 376) = 1.931, \textit{p} = .040 (see Table~\ref{tab:posthoc}). For text-based MCQs, ReQUESTA MCQs differed from GPT-5 MCQs, Pillai's Trace = .062, F(5, 187) = 2.455, \textit{p} = .035, partial $\eta^2$ = .062. Specifically, ReQUESTA MCQs received higher scores for topic relevance than GPT-5, \textit{p} = .010. For inferential MCQs, ReQUESTA MCQs differed from GPT-5 MCQs, Pillai's Trace = .159, F(5, 187) = 7.048, \textit{p} < .001, partial $\eta^2$ = .159. Specifically, ReQUESTA outperformed GPT-5 on distractor linguistic features and distractor semantic plausibility, \textit{p}s < .05, whereas GPT-5 outperformed ReQUESTA on writing clarity, \textit{p} < .001. For main idea MCQs, ReQUESTA MCQs differed from GPT-5 MCQs, Pillai's Trace = .092, F(5, 187) = 3.768, \textit{p} = .003, partial $\eta^2$ = .092. Specifically, ReQUESTA produced higher scores for distractor semantic plausibility, \textit{p} = .002, and GPT-5 received higher scores for writing clarity, \textit{p} = .011. All other comparisons were nonsignificant.

\begin{table}[t]
\caption{Post-hoc analyses of the effect of question source and question type on question quality scores. Each $F$ tests the simple effects of question source within each level of question type.}
\label{tab:posthoc}
\centering
\begin{tabular}{|l|l|c|c|}
\hline
\textbf{Scoring Category} & \textbf{Question Type} & \textbf{\textit{F}(1, 191)} & \textbf{$\eta^2$} \\
\hline
\multirow{3}{*}{Topic Relevance} 
& Text-Based & 6.75 & 0.03* \\ 
& Inferential & 2.70 & 0.01 \\
& Main Idea & 0.00 & 0.00 \\
\hline
\multirow{3}{*}{Writing Clarity} 
& Text-Based & 0.05 & 0.00 \\ 
& Inferential & 14.90 & 0.07*** \\
& Main Idea & 6.64 & 0.03* \\
\hline
\multirow{3}{*}{Distractor Linguistic Features} 
& Text-Based & 1.96 & 0.01 \\ 
& Inferential & 7.04 & 0.04** \\
& Main Idea & 0.22 & 0.00 \\
\hline
\multirow{3}{*}{Distractor Semantic Plausibility} 
& Text-Based & 1.52 & 0.01 \\ 
& Inferential & 5.01 & 0.03* \\
& Main Idea & 9.78 & 0.05** \\
\hline
\multirow{3}{*}{Distractor Semantic Uniqueness} 
& Text-Based & 0.25 & 0.00 \\ 
& Inferential & 1.90 & 0.01 \\
& Main Idea & 0.41 & 0.00 \\
\hline
\end{tabular}

\vspace{2mm}
\noindent\textit{Note.} *\textit{p} < .05. **\textit{p} < .01. ***\textit{p} < .001.
\end{table}

\section{Discussion}

This study demonstrates that workflow-level, agentic orchestration can systematically improve the quality of automatically generated MCQs through system design, independent of changes to the underlying large language model. Across complementary evaluation layers, including psychometric analyses based on learner performance and expert judgments of item quality, ReQUESTA consistently outperformed a single-pass GPT-5 baseline. These findings suggest that progress in automated assessment generation depends not only on increasingly capable language models, but critically on deliberate system-level design choices that govern how those models are prompted, constrained, evaluated, and iteratively refined. In this sense, ReQUESTA functions as an engineering pattern for transforming a general-purpose LLM into a more reliable generator of structured artifacts under multiple simultaneous constraints.

\subsection{Psychometric Evidence of Cognitive Rigor}
Our first research question centered on the difference in psychometric properties between MCQs generated by ReQUESTA and those generated by GPT-5. The results from CTT analyses indicate that ReQUESTA-generated items were, on average, more cognitively challenging, more discriminative, and more strongly aligned with overall reading comprehension performance than items generated by the GPT-5 baseline. Mixed-effects logistic regression analyses further corroborated this pattern, showing that ReQUESTA items were associated with significantly lower probabilities of correct responses even after accounting for question type and individual differences in vocabulary knowledge. Importantly, this effect was consistent across text-based, inferential, and main idea questions, indicating that the framework’s advantages were not confined to a single cognitive category.

From an AI engineering perspective, these results suggest that plan- and constraint-guided generation can systematically change the functional behavior of LLM-generated artifacts. The ReQUESTA pipeline introduces explicit intermediate planning representations, assigns distinct generation roles to different agents, couples probabilistic LLM outputs with rule-based controllers that enforce task boundaries, and applies automated evaluation and revision cycles. Together, these mechanisms shift generation away from opportunistic, single-pass text production toward more controlled construction of items with specific reasoning requirements. The observed changes in difficulty and discrimination thus reflect not just surface differences, but measurable effects of workflow-level control over what cognitive operations are required to answer the questions correctly. 

\subsection{Expert Judgments of Item and Distractor Quality}
Our second research question focused on the differences in expert evaluations between ReQUESTA and ChatGPT 5 generated MCQs. Analysis of the expert evaluations provides converging evidence for the advantages of ReQUESTA at the level of item construction. Compared with the GPT-5 baseline, ReQUESTA-generated questions were rated as more topic-relevant, indicating stronger alignment with central ideas in the source texts rather than peripheral or trivial details. Because the pedagogical value of MCQ-based assessments depends critically on whether items emphasize core concepts, this finding is particularly important. Prior research has shown that AI-generated questions often overemphasize isolated facts or superficial details \cite{ling2024automatic}; ReQUESTA appears to mitigate this tendency by explicitly planning question generation around key concepts and overarching themes.

ReQUESTA also demonstrated clear advantages in distractor quality. Experts rated ReQUESTA distractors as more linguistically consistent in surface features such as length and syntactic structure, as well as more semantically plausible to learners who had not fully comprehended the text. These improvements were especially pronounced for inferential questions, which are widely recognized as one of the most challenging components of automated MCQ generation. Whereas previous studies have documented that AI-generated MCQs often include distractors that are stylistically inconsistent, implausible, or easily eliminated due to obvious semantic mismatches \cite{bitew2023distractor, yu2024enhancing}, ReQUESTA produced distractors that more closely resembled those found in carefully authored human assessments. 

At the same time, expert ratings revealed a trade-off between cognitive complexity and perceived clarity. As questions increased in cognitive demand, particularly for inferential and main idea items, experts judged some ReQUESTA-generated questions to be less clear or concise. This pattern likely reflects an inherent tension in assessment design: questions that require integration, abstraction, or synthesis often necessitate richer contextual framing, which can challenge strict standards of brevity and surface clarity. This trade-off highlights an important design consideration for automated assessment systems and underscores the need to balance cognitive rigor with readability and presentation quality.

\subsection{Cognitive Alignment Through Agentic Design}
Taken together, the psychometric and expert evidence indicates that cognitive targeting emerges from orchestration and workflow design, not merely from scaling model capacity. Single-pass prompting delegates planning, constraint satisfaction, and quality assurance to the latent behavior of a single model call \cite{kojima2022large}. In contrast, ReQUESTA externalizes these functions into a coordinated workflow that (a) separates planning from execution, (b) assigns specialized roles to generation subtasks, and (c) incorporates iterative evaluation and revision. The observed improvements are therefore best understood as the product of decomposition plus verification loops, aligning with broader trends in agentic LLM systems where reliability is improved by structuring tasks into auditable intermediate steps and enforcing constraints through repeated checks \cite{christie2025agentic, lu2025octotools}.

A second implication concerns modularity and maintainability. Because ReQUESTA is composed of separable agents, individual agents or modules can be modified, replaced, or augmented without redesigning the entire system, enabling the integration of new language models, external tools, or evaluation criteria as they become available. This modularity allows the framework to accommodate the strengths and limitations of different models across subtasks and supports adaptation to diverse contexts and objectives. More generally, ReQUESTA illustrates a hybrid design pattern for controllable structured generation: keep the base model constant, allocate generative flexibility to LLM agents, delegate enforcement and coordination to rule-based agents, and validate improvements using outcome measures that reflect artifact function (e.g., discrimination, plausibility, and constraint satisfaction) rather than surface  quality alone.

\section{Conclusion}
This study introduced ReQUESTA, a hybrid, multi-agent framework for automated MCQ generation designed to support cognitively diverse assessment at scale. By decomposing the MCQ authoring process into specialized subtasks and orchestrating LLMs with rule-based control and evaluation mechanisms, ReQUESTA enables the systematic generation of text-based, inferential, and main idea questions from academic texts. Through a combination of psychometric analyses based on learner responses and structured expert evaluations of item quality, the study demonstrated that ReQUESTA consistently outperformed a single-pass GPT-5 baseline using the same underlying language model. In particular, ReQUESTA-generated items were more challenging, more discriminative, and more strongly aligned with central concepts in the source texts, while also exhibiting higher-quality distractors in terms of linguistic consistency and semantic plausibility.

Taken together, these findings make an important contribution to the literature on automated assessment generation. They provide empirical evidence that improvements in MCQ quality are not solely driven by increasingly capable language models, but can be achieved through deliberate system-level design that emphasizes agentic decomposition, structured prompting, constraint enforcement, and iterative evaluation. From an AI engineering perspective, ReQUESTA illustrates how agentic orchestration can be used to shape the functional properties of generated artifacts, enabling more reliable control over task demands and output structure than is typically achievable with single-pass prompting.

\subsection{Limitations}
Several limitations of the present study should be acknowledged. First, the empirical evaluation focused on academic expository texts drawn from a limited set of OpenStax textbook domains, and passages were constrained to approximately 400 words to balance topic diversity and participant fatigue. Although this design supports controlled comparison, texts from other disciplines, genres, or with substantially different lengths may interact differently with the framework and influence relative performance compared to a single-pass baseline. Second, ReQUESTA operationalized cognitive diversity using three question categories (text-based, inferential, and main idea). While these categories are well-established and tractable for evaluation, they do not exhaust the space of possible task demands that could be targeted in structured generation systems. These considerations suggest caution in generalizing the findings beyond the specific contexts examined in this study.

\subsection{Future Work}
Future work will extend both the technical capabilities and empirical scope of ReQUESTA. One immediate direction is the development of a more robust automated MCQ evaluation component that is explicitly validated against expert judgments, replacing the current evaluator that relies primarily on checklist-based prompting. Such an evaluation module would enable more reliable quality monitoring and finer-grained control during question generation. In addition, future iterations of the framework will expand the range of targeted cognitive categories to include application- and evaluation-level questions, allowing ReQUESTA to address a broader spectrum of learning objectives. Finally, broader empirical validation across additional domains, text types, and populations will be pursued to assess the generalizability of agentic orchestration strategies for structured generation tasks beyond MCQ authoring.

\section*{Acknowledgments}

The research reported here was supported by Arizona State University (ASU), the ASU Learning Engineering Institute, and the Institute of Education Sciences, U.S.\ Department of Education, through Grant R305T240035 to Arizona State University. The opinions expressed are those of the authors and do not represent the views of Arizona State University, the Institute of Education Sciences, or the U.S.\ Department of Education.

\bibliographystyle{splncs04}
\bibliography{sigproc}  
%

%
%
%

\clearpage
\section{Appendix}

\subsubsection{Expert Evaluation Rubric}
\noindent\mbox{} 
\FloatBarrier   

\begin{table}[ht]
\centering
\caption{Binary evaluation criteria (0 = Disagree, 1 = Agree)}
\label{tab:binary-eval}
\renewcommand{\arraystretch}{1.4}

\begin{tabular}{|
>{\raggedright\arraybackslash}p{0.50\textwidth}|
>{\centering\arraybackslash}p{0.22\textwidth}|
>{\centering\arraybackslash}p{0.22\textwidth}|}
\hline
 & \textbf{0 -- Disagree} & \textbf{1 -- Agree} \\
\hline

\textbf{Answer Correctness}

\vspace{3pt}
Correct answer is clearly correct to those who have comprehended the text.
&  &  \\
\hline

\textbf{Distractor Incorrectness}

\vspace{3pt}
Distractors are clearly incorrect to those who have comprehended the text.
&  &  \\
\hline
\end{tabular}
\end{table}

\clearpage  

\begin{table}[ht]
\centering
\caption{Four-point rubric for item and distractor quality}
\label{tab:rubric-quality}
\small
\begin{tabularx}{\textwidth}{@{}
>{\raggedright\arraybackslash}p{0.24\textwidth}
>{\raggedright\arraybackslash}X
>{\raggedright\arraybackslash}X
>{\raggedright\arraybackslash}X
>{\raggedright\arraybackslash}X
@{}}
\toprule
\textbf{Criterion} &
\textbf{1 -- Disagree} &
\textbf{2 -- Neutral} &
\textbf{3 -- Agree} &
\textbf{4 -- Strongly Agree} \\
\midrule

Overall quality &
The question and options are unclear, ambiguous, or verbose. &
Some clarity issues and/or minor ambiguity remain. &
Clearly worded, concise, and unambiguous. &
Excellent clarity: concise, unambiguous, and professional. \\

\addlinespace
Topic relevance &
Does not address a central concept in the text. &
Addresses a peripheral or partially relevant point. &
Addresses a central concept in the text. &
Addresses a central concept very directly and importantly. \\

\addlinespace
Coverage (where content appears) &
Not covered in main points or sub-points. &
Partially covered in sub-points or only partially covered overall. &
Partially covered in a main point; fully covered in sub-points. &
Fully covered as a main point in the ideas sheet. \\

\addlinespace
Writing clarity (stem \& options) &
Stem and options are excessively wordy or unclear. &
Wordiness or mild ambiguity that slightly impacts comprehension. &
Stem and options concise with minor redundancies. &
Stem and options fully clear and concise. \\

\addlinespace
Distractor plausibility &
Distractors are implausible or obviously wrong. &
Some distractors somewhat implausible or uneven. &
Distractors plausible to those who have not comprehended the text. &
All distractors are highly plausible and well matched. \\

\addlinespace
Linguistic features (length / surface similarity) &
Distractors visibly different from the correct answer (e.g., much longer or shorter). &
Some distractors have noticeable differences or one major difference. &
Distractors have relatively similar length and surface features. &
All distractors are uniform in linguistic features and closely matched. \\

\bottomrule
\end{tabularx}
\end{table}

\clearpage
\section*{\centering Supplementary Materials}

\subsection*{\centering Representative MCQs Generated by ReQUESTA and the GPT-5 Zero-Shot Baseline}
\vspace{12pt}
\subsubsection*{Passage One (Anthropology)}\leavevmode\\
Biological anthropology focuses on the earliest processes in the biological and sociocultural development of human beings as well as the biological diversity of contemporary humans. In other words, biological anthropologists study the origins, evolution, and diversity of our species. Some biological anthropologists use genetic data to explore the global distribution of human traits such as blood type or the ability to digest dairy products. Some study fossils to learn how humans have evolved and migrated. Some study our closest animal relatives, the primates, in order to understand what biological and social traits humans share with primates and explore what makes humans unique in the animal world.

The Dutch primatologist Carel van Schaik spent six years observing orangutans in Sumatra, discovering that these reclusive animals are actually much more social than previously thought (2004). Moreover, van Schaik observed that orangutans use a wide variety of tools and pass down skills to their young. By studying these primates, van Schaik and other biological anthropologists gain insight into the origins of human intelligence, technology, and culture. These researchers also warn that habitat loss, illegal hunting, and the exotic pet trade threaten the survival of our fascinating primate cousins.

Biological anthropologists frequently combine research among primates with evidence from the human fossil record, genetics, neuroscience, and geography to answer questions about human evolution. Sometimes their insights are startling and unexpected. Anthropologist Lynne Isbell argues that snakes have played a key role in the evolution of human biology, particularly our keen sense of sight and our ability to communicate through language (Isabell, 2009). Isbell's "snake detection theory" posits that primates developed specialized visual perception as well as the ability to communicate what they were seeing in order to alert others to the threat of venomous snakes in their environment. She points to the near-universal fear of snakes shared by both humans and primates and has documented the prevalence of snake phobia in human myth and folklore. Isbell's research highlights how human-animal relations are central to humanity, shaping both biology and culture.

Not all biological anthropologists study primates. Many biological anthropologists study fossilized remains in order to chart the evolution of early hominins, the evolutionary ancestors of modern humans. In this field of study, anthropologists consider the emergence and migration of the various species in the hominin family tree as well as the conditions that promoted certain biological and cultural traits. Some biological anthropologists examine the genetic makeup of contemporary humans in order to learn how certain genes and traits are distributed in human populations across different environments. Others examine human genetics looking for clues about the relationships between early modern humans and other hominins, such as Neanderthals.

\begin{requestabox}
\begin{center}
\textbf{ReQUESTA-Generated MCQs}
\end{center}

\qtype{Text-Based}

Which combination of research approaches is used in biological anthropology to investigate human evolution?

\begin{enumerate}[label=\Alph*)]
\item Fossil record, genetics, linguistics, archaeology, and geography.
\item Fossil record, genetic data, primate studies, neuroscience, and geography.
\item Genetics, primate studies, archaeology, ecology, and linguistics.
\item Neuroscience, geography, cultural ethnography, economics, and political science.
\end{enumerate}
\answer{B}
\qdotted

\qtype{Text-Based}
What adaptations does Isbell's snake detection theory propose primates evolved in response to snake threats?

\begin{enumerate}[label=\Alph*)]
\item Greater manual skill for tool use to capture snakes.
\item Heightened smell and nocturnal habits to avoid predators.
\item Larger teeth and threat displays to deter reptile attacks.
\item Refined vision and communication to warn of venomous snakes.
\end{enumerate}
\answer{D}
\qdotted

\qtype{Inferential}

Which inference is best supported by van Schaik's findings that orangutans use tools and transmit skills to their young?

\begin{enumerate}[label=\Alph*)]
\item Cultural learning and problem-solving likely reflect capacities that predate humans and extend across primates.
\item Human cultural distinctiveness arises chiefly from language shaped by ancestral snake-related threats.
\item Orangutan societies exhibit cognition and traditions essentially on par with those of humans.
\item Their tool behaviors are probably products of captivity or training rather than natural development.
\end{enumerate}
\answer{A}
\qdotted

\qtype{Inferential}

If the pressures of habitat loss, illegal hunting, and the pet trade persist, what is the most likely impact on research into human evolution?

\begin{enumerate}[label=\Alph*)]
\item A pivot to fossil evidence that would largely supplant insights from observing living species.
\item Diminished access to living comparative models, constraining inferences about human origins.
\item Expanded testing of the snake-detection theory, compensating for the loss of field observations of apes.
\item Greater emphasis on genetic surveys of contemporary populations, yielding similar clarity about early evolution.
\end{enumerate}
\answer{B}
\qdotted

\qtype{Main Idea}

Which statement best summarizes the text?

\begin{enumerate}[label=\Alph*)]
\item Biological anthropology integrates evidence from primates, fossils, and genetics to explain human origins, evolution, and diversity.
\item Primate research revises earlier assumptions about orangutan social behavior.
\item Fossil and genetic studies chart trait distributions across populations.
\item Snake detection theory explains fear and folklore.
\end{enumerate}
\answer{A}

\end{requestabox}

\begin{baselinebox}
\begin{center}
\textbf{GPT-5 Zero-Shot Baseline MCQs}
\end{center}

\qtype{Text-Based}

What did Carel van Schaik discover about orangutans during his study in Sumatra?

\begin{enumerate}[label=\Alph*)]
\item They prefer solitude and avoid social interactions.
\item They are less intelligent than previously thought.
\item They are more social and use a wide variety of tools.
\item They have no ability to learn new skills.
\end{enumerate}
\answer{C}
\qdotted

\qtype{Text-Based}

According to Lynne Isbell, what role did snakes play in human evolution?

\begin{enumerate}[label=\Alph*)]
\item They were a primary food source for early humans.
\item They contributed to the development of human language and sight.
\item They had no significant impact on human evolution.
\item They caused the extinction of early hominins.
\end{enumerate}
\answer{B}
\qdotted

\qtype{Inferential}

Why might biological anthropologists study the genetic makeup of contemporary humans?

\begin{enumerate}[label=\Alph*)]
\item To find ways to improve human genetic traits.
\item To create a universal human genetic profile.
\item To understand ancient human diets.
\item To learn how certain genes and traits are distributed across environments.
\end{enumerate}
\answer{D}
\qdotted

\qtype{Inferential}

What might be a consequence of habitat loss and illegal hunting for primates, according to the text?

\begin{enumerate}[label=\Alph*)]
\item Primates will become more social.
\item Human intelligence will decline.
\item The survival of primates is threatened.
\item Primates will migrate to urban areas.
\end{enumerate}
\answer{C}
\qdotted

\qtype{Main Idea}

What is the main focus of the text on biological anthropology?

\begin{enumerate}[label=\Alph*)]
\item The study of fossils and ancient human remains exclusively.
\item The impact of modern technology on human evolution.
\item The exploration of human origins, evolution, and biological diversity through various methods.
\item Primatology as the only significant area of study in anthropology.
\end{enumerate}
\answer{C}

\end{baselinebox}

\vspace{12pt}
\subsubsection*{Passage Two (Sociology)}\leavevmode\\

Although human societies have much in common, cultural differences are far more prevalent than cultural universals. For example, while all cultures have language, analysis of conversational etiquette reveals tremendous differences. In some Middle Eastern cultures, it is common to stand close to others in conversation. Americans keep more distance and maintain a large "personal space." Additionally, behaviors as simple as eating and drinking vary greatly from culture to culture. Some cultures use tools to put the food in the mouth while others use their fingers. If your professor comes into an early morning class holding a mug of liquid, what do you assume they are drinking? In the U.S., it's most likely filled with coffee, not Earl Grey tea, a favorite in England, or Yak Butter tea, a staple in Tibet.

Often, however, people express disgust at another culture's cuisine. They might think that it's gross to eat raw meat from a donkey or parts of a rodent, while they don't question their own habit of eating cows or pigs.

Such attitudes are examples of ethnocentrism, which means to evaluate and judge another culture based on one's own cultural norms. Ethnocentrism is believing your group is the correct measuring standard and if other cultures do not measure up to it, they are wrong. As sociologist William Graham Sumner (1906) described the term, it is a belief or attitude that one's own culture is better than all others. Almost everyone is a little bit ethnocentric.

A high level of appreciation for one's own culture can be healthy. A shared sense of community pride, for example, connects people in a society. But ethnocentrism can lead to disdain or dislike of other cultures and could cause misunderstanding, stereotyping, and conflict. Cultural imperialism is the deliberate imposition of one's own cultural values on another culture. 

Colonial expansion by Portugal, Spain, Netherlands, and England grew quickly in the fifteenth century was accompanied by severe cultural imperialism. European colonizers often viewed the people in these new lands as uncultured savages who needed to adopt Catholic governance, Christianity, European dress, and other cultural practices.

A modern example of cultural imperialism may include the work of international aid agencies who introduce agricultural methods and plant species from developed countries into areas that are better served by indigenous varieties and agricultural approaches to the particular region. 
Another example would be the deforestation of the Amazon Basin as indigenous cultures lose land to timber corporations.

\begin{requestabox}
\begin{center}
\textbf{ReQUESTA-Generated MCQs}
\end{center}

\qtype{Text-Based}

What does ethnocentrism mean?

\begin{enumerate}[label=\Alph*)]
\item Deliberately imposing one group's values and practices on another.
\item Expressing pride in one's community to strengthen social bonds.
\item Judging other cultures by one's standards, deeming yours superior.
\item Recognizing common features shared by societies, such as language.
\end{enumerate}
\answer{C}
\qdotted

\qtype{Text-Based}

What is cultural imperialism?

\begin{enumerate}[label=\Alph*)]
\item Assessing other cultures using one's own cultural standards.
\item Common features present across all societies.
\item Deliberately imposing one group's values on another society.
\item Fostering social cohesion through pride in communal heritage.
\end{enumerate}
\answer{C}
\qdotted

\qtype{Inferential}

Given that many people judge other cultures by their own standards and this can produce misunderstanding, stereotyping, and conflict, which approach would most likely enhance effective cross-cultural interactions?

\begin{enumerate}[label=\Alph*)]
\item Emphasizing pride in one's heritage while downplaying differences with other groups.
\item Establishing uniform behavioral standards to align practices across diverse societies.
\item Minimizing contact with unfamiliar groups to reduce clashes over customs and beliefs.
\item Promoting awareness of differing customs and questioning assumptions about unfamiliar practices.
\end{enumerate}
\answer{D}
\qdotted

\qtype{Inferential}

Why might the introduction of foreign farming methods by aid groups and the expansion of timber companies into the Amazon both exemplify cultural imperialism?

\begin{enumerate}[label=\Alph*)]
\item Both arise because all cultures share the same basic needs and therefore benefit from uniform solutions.
\item Both prioritize outside agendas that displace region-specific knowledge and erode indigenous lands and livelihoods.
\item Both reflect the natural diffusion of culture through voluntary exchange.
\item Both spread beneficial innovations that raise productivity without affecting local customs.
\end{enumerate}
\answer{B}
\qdotted

\qtype{Main Idea}

Which statement best summarizes the text?

\begin{enumerate}[label=\Alph*)]
\item Cultural differences outweigh universals, as seen in conversational distance and eating and drinking practices.
\item Ethnocentrism is a healthy expression of pride that generally strengthens societies and reduces conflict.
\item Imposing one culture's values on another is mainly a feature of European colonialism and has limited relevance today.
\item Wide cultural variation is common, and judging or imposing one culture's norms (ethnocentrism, cultural imperialism) causes harm.
\end{enumerate}
\answer{D}

\end{requestabox}

\begin{baselinebox}
\begin{center}
\textbf{GPT-5 Zero-Shot Baseline MCQs}
\end{center}

\qtype{Text-Based}

What is an example of a cultural universal mentioned in the text?

\begin{enumerate}[label=\Alph*)]
\item Eating with utensils
\item Use of language
\item Drinking coffee
\item Wearing European dress
\end{enumerate}
\answer{B}
\qdotted

\qtype{Text-Based}

According to the text, what is cultural imperialism?

\begin{enumerate}[label=\Alph*)]
\item The appreciation of one's own culture
\item The imposition of one's own cultural values on another culture
\item The belief that all cultures are equal
\item The practice of respecting indigenous cultures
\end{enumerate}
\answer{B}
\qdotted

\qtype{Inferential}

Based on the text, what is a potential consequence of international aid agencies introducing foreign agricultural methods?

\begin{enumerate}[label=\Alph*)]
\item Improved local agricultural productivity
\item Loss of biodiversity and local agricultural practices
\item Increased cultural diversity in the region
\item Strengthening of indigenous cultural practices
\end{enumerate}
\answer{B}
\qdotted

\qtype{Inferential}

What can be inferred about ethnocentrism from the text?

\begin{enumerate}[label=\Alph*)]
\item It encourages understanding and acceptance of different cultures
\item It can lead to negative outcomes such as conflict and stereotyping
\item It is a modern phenomenon that did not exist in the past
\item It is primarily a positive force within societies
\end{enumerate}
\answer{B}
\qdotted

\qtype{Main Idea}

What is the main idea of the text?

\begin{enumerate}[label=\Alph*)]
\item Cultural universals are more common than cultural differences
\item Ethnocentrism is beneficial for maintaining cultural identity
\item Cultural differences and ethnocentrism can lead to cultural imperialism
\item All cultures should adopt Western cultural practices
\end{enumerate}
\answer{C}

\end{baselinebox}

\end{document}